\documentclass[conference]{IEEEtran}
\IEEEoverridecommandlockouts
\usepackage{cite}
\usepackage{amsmath,amssymb,amsfonts}
\usepackage{algorithmic}
\usepackage{graphicx}
\usepackage{textcomp}
\usepackage{xcolor}

\usepackage{lipsum}
\usepackage{graphicx}
\ifCLASSOPTIONcompsoc
    \usepackage[caption=false, font=normalsize, labelfont=sf, textfont=sf]{subfig}
\else
\usepackage[caption=false, font=footnotesize]{subfig}
\fi

\def\BibTeX{{\rm B\kern-.05em{\sc i\kern-.025em b}\kern-.08em
    T\kern-.1667em\lower.7ex\hbox{E}\kern-.125emX}}
\begin{document}
\begin{table*}
IEEE Copyright Notice

© 2019 IEEE. Personal use of this material is permitted. Permission from IEEE
must be obtained for all other uses, in any current or future media, including
reprinting/republishing this material for advertising or promotional purposes,
creating new collective works, for resale or redistribution to servers or lists, or reuse
of any copyrighted component of this work in other works.

\textbf{Published by:}
MVIP 2020 - The 11th Iranian and the first International Conference on Machine Vision and Image Processing. Qom , Iran.

\textbf{DIO:} 10.1109/MVIP49855.2020.9116870
\end{table*}

\title{An Efficient Approach for Using Expectation Maximization Algorithm in Capsule Networks\\}

\author{
\IEEEauthorblockN{Moein Hasani}
\IEEEauthorblockA{\textit{Department of Computer Engineering} \\
\textit{Bu Ali Sina University}\\
Hamedan, Iran \\
m.hasani@eng.basu.ac.ir}

\and

\IEEEauthorblockN{Amin Nasim Saravi}
\IEEEauthorblockA{\textit{Department of Computer Engineering} \\
\textit{Bu Ali Sina University}\\
Hamedan, Iran \\
a.nasimsaravi@eng.basu.ac.ir}

\and

\IEEEauthorblockN{Hassan Khotanlou}
\IEEEauthorblockA{\textit{Department of Computer Engineering} \\
\textit{Bu Ali Sina University}\\
Hamedan, Iran \\
khotanlou@basu.ac.ir}
}

\maketitle

\begin{abstract}
Capsule Networks (CapsNets) are brand-new architectures that have shown ground-breaking results in certain areas of Computer Vision (CV). In 2017, Hinton and his team introduced CapsNets with routing-by-agreement in “Sabour et al” and in a more recent paper “Matrix Capsules with EM Routing” they proposed a more complete architecture with Expectation-Maximization (EM) algorithm. Unlike the traditional convolutional neural networks (CNNs), this architecture is able to preserve the pose of the objects in the picture. Due to this characteristic, it has been able to beat the previous state-of-the-art results on the smallNORB dataset, which includes images with various view points. Also, this new architecture is more robust to white box adversarial attacks. However, CapsNets have two major drawbacks. They can’t perform as well as CNNs on complex datasets and, they need a huge amount of time for training. We try to mitigate these shortcomings by finding optimum settings of EM routing iterations for training CapsNets. Unlike the past studies, we use un-equal numbers of EM routing iterations for different stages of the CapsNet. We manage to achieve higher accuracies than the original CapsNet while training the network up to three times faster. For our research, we use three datasets: Yale face dataset, Belgium Traffic Sign dataset, and Fashion-MNIST dataset.
\end{abstract}


\section{Introduction}
During the last decade, Deep Learning has contributed to the success of computer vision enormously. The dominant models used in Computer Vision (CV) are Convolutional Neural Networks (CNNs). These networks have achieved extraordinary results in different tasks of CV. However, there are some problems with these networks. The convolution operation and pooling operation in CNNs cause the input to be down-sampled, which leads to losing some valuable information. Furthermore, CNNs are vulnerable to adversarial attacks. 
Capsule Networks (CapsNets) are new architectures that have shown ground-breaking results in certain areas of computer vision. Geoffry Hinton and his team, first introduced the concept of CapsNets in 2011, in a paper titled “Transforming Autoencoders” \cite{b6}. In 2017, Hinton and his team introduced dynamic routing between capsules in “Sabour et al” \cite{b5} to make capsules more practical. In a more recent paper “Matrix Capsules with EM Routing” \cite{b18} they proposed a more complete model that works with matrices and benefits from the Expectation-Maximization (EM) algorithm \cite{b17}. Unlike the traditional CNNs, this architecture is able to preserve the pose of the objects in the picture. Due to this characteristic, it has been able to beat the previous state-of-the-art results on the smallNORB dataset\cite{b16}, reducing the number of errors by 45\%, and it is more robust to white box adversarial attacks. 
Even though the CapsNets outperform traditional CNNs in some scenarios, they have two major drawbacks. First, they can’t perform as well as CNNs on complex datasets like CIFAR10 \cite{b8}. Second, they need a huge amount of time to be trained because of the routing-by-agreement algorithm. Due to these shortcomings, there are not so many research papers on the CapsNets. 
We plan to accelerate the training process of the CapsNets while reaching higher accuracies. We study the effects that different numbers of EM routing iterations have on the performance of the CapsNets. For our experiments, we evaluate the performance of CapsNet on three datasets with different themes. We allow our networks to converge on these datasets in order to evaluate the highest accuracy that can be achieved.

\section{Matrix Capsule Network with EM Routing}
CapsNets are made of layers of capsules and each capsule is a group of neurons. Matrix capsule is the last version of the capsule networks introduced by the author team. A matrix capsule, similar to a neuron, captures the likeliness of occurring an object in a picture. It can also capture the pose information related to the object and stores it in a 4x4 matrix. Each capsule in the lower layer (child capsule) makes predictions (votes) on the pose matrices of the capsules in the next layer (parent capsule). Child capsules with similar votes are assigned to the same parent capsule. Similar votes mean that the clustered capsules agree on the pose matrix of the parent capsule. Each vote is computed by multiplying the pose matrix of a child capsule with the viewpoint invariant transformation (VIT) matrix. The VIT matrix is learned through the backpropagation process. To group capsules to form the part-whole (child-parent) relationship, the EM routing algorithm is used. The EM algorithm starts with random initialization of clusters (Gaussian distributions) and tries to fit the training data-points (votes) into the clusters. After that, it re-computes the mean and the standard deviation for each cluster with respect to the data-points assigned to them. For training CapsNets, the mentioned process is performed in two parts; The E-step which determines the assignment probability of each child capsule to a parent capsule and the M-step which re-calculates the values of Gaussian distributions (mean and standard deviation) based on the assignment probabilities. By each iteration of this procedure, we try to converge to Gaussian distributions that maximize the likelihood of the observed data-points in order to activate the right parent capsule.

\section{Methods}
\subsection{Network Architecture}\label{NA}
We use the same architecture as the original capsule network paper. The architecture is depicted in Fig. 1. This architecture includes five main stages: ReLUConv, PrimaryCaps, ConvCaps1, ConvCaps2, and ClassCaps. Each stage uses a number of Convolution (Conv) filters with different sizes and strides. Our choice for the number of the filters and their sizes is based on the most efficient setting reported in previous experiments \cite{b14,b15,b18}. This setting for the Conv filters is known to work efficiently. The outputs of ConvCaps1, ConvCaps2, and ClassCaps are computed using EM routing. ReLUConv is a regular Conv layer with ReLU \cite{b4} activation function. We use 64 filters of size 5x5 with a stride of 2 which output 64 (A = 64) channels. In PrimaryCaps, we employ a 1x1 Conv filter to transform the 64 channels from the ReLUConv stage into 8 (B = 8) primary capsules. Each capsule contains a 4x4 pose matrix and an activation value. ConvCaps1 comes after the PrimaryCaps and likewise, it outputs capsules. In this stage, we use 3x3 filters (K=3) with a stride of 2 to produce the output capsules. The capsule outputs of ConvCaps1 are then fed into ConvCaps2 which again uses the filters with a window size of 3x3 but with a stride of 1. The output capsules of ConvCaps2 are connected to the ClassCaps which utilizes 1x1 filters and outputs one capsule per class.
\begin{figure}[b]
\centerline{\includegraphics[width=0.5\textwidth]{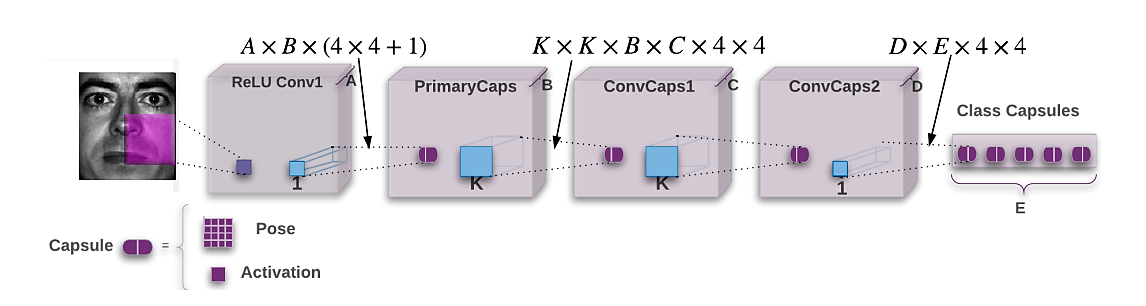}}
\caption{Architecture of the CapsNet we used. In our work: A=64, B=8, C=16, D=16, and K=3. (The image is from the original paper.)}
\label{arch}
\end{figure}
\subsection{Different Routings}
Previous experiments on the CapsNets have always practiced using the same number of routing iterations for stages that benefit from EM routing. In contrast, we select a specific number of iterations for each stage in our study. The results of previous works have demonstrated that increasing the number of iterations of the EM routing algorithm can be beneficial to the network’s ability to learn but, it starts to hurt the performance after reaching a certain number \cite{b9,b14}. As a result, we avoid using large numbers for EM routing iterations. In our reports, we use a three-digit representation where each digit in this representation shows the number of iterations the EM routing algorithm is run for each of the ConvCaps1, ConvCaps2, and ClassCaps stages respectively. We reference these digits as d1, d2, and d3.

\subsection{Loss Function}
Based on \cite{b18}, we have picked the spread loss as the main loss function for optimizing our networks. The loss for a wrong class i is defined as:
\begin{equation}
L_{i}=(\max(0,m-(a_{t}-a_{i}))^{2}\label{losseq}
\end{equation}
Where $a_{t}$ is the activation of the true class and $a_{i}$ is the activation predicted by the network for wrong classes. The total cost for a sample is shown in:
\begin{equation}
L=\sum_{i\neq t}L_{i}\label{totallosseq}
\end{equation}
If the margin between the true label and the wrong class is smaller than m, it will be penalized by the square of $m-(a_{t}-a_{i})$. The initial value of m is 0.2 and it is linearly increased by 0.1 after each training epoch until reaching a maximum of 0.9.

\section{Experiments and Results}

\begin{figure*}
\centering
\subfloat{%
\includegraphics[width=0.33\linewidth]{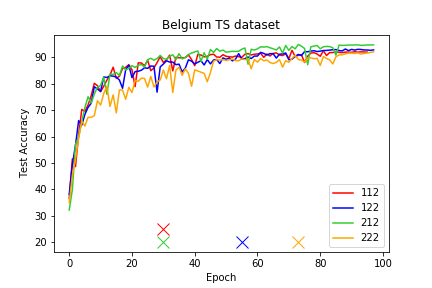}}
\subfloat{%
\includegraphics[width=0.33\linewidth]{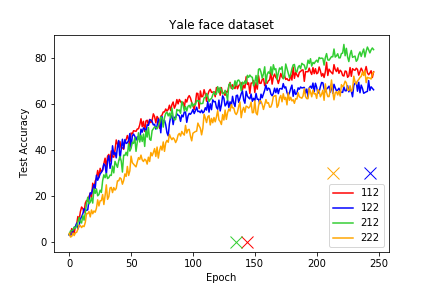}}
\subfloat{%
\includegraphics[width=0.33\linewidth]{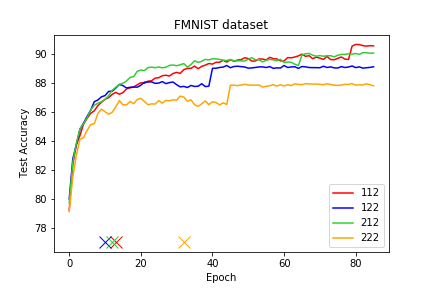}}
\caption{Training process of our proposed models on the datasets. The X marks show when each model reaches the selected threshold. The thresholds for Belgium TS dataset, Yale Face dataset, and FMNIST dataset are 90\%, 70\%, 87\% respectively.}
\end{figure*}

\subsection{Setup}
We adopt an Nvidia GeForce 1070 Ti, 16 GB of RAM, and an Intel Core i7 extreme 7\textsuperscript{th} generation CPU for carrying out the training of the models. The PyTorch library \cite{b10} has been used for implementing the models.
To evaluate the performance of our different models, we use three datasets: Yale face dataset B \cite{b1}, Belgium Traffic Sign (TS) dataset \cite{b11}, and Fashion-MNIST (FMNIST) dataset \cite{b12}. These datasets are in three different areas of CV which can help us evaluate our proposed way of training the CapsNet better.
\\Belgium TS dataset encompasses 7000 images of 62 classes of traffic signs from Belgium. The images in this dataset have diverse illumination levels and points of view. There are several impediments including dirt, stickers, and trees that add occlusion to the images of this dataset which makes this dataset a good candidate for evaluating the performance of the CapsNet. We transform the images from RGB to grey-scale and resize all the images from different sizes to 40x40. Random rotation is applied to the pictures of this dataset as an augmentation technique.
\\Yale face dataset B is consisting of grayscale images in 38 classes. Pictures in this data set have varying illumination conditions and points of view which makes us able to asses the CapsNets in challenging lighting conditions. We use a subset of size 2452 of this dataset. The images are center cropped and, we resize all of them to 40x40. Random rotation and random horizontal flip are our selected augmentation techniques for this dataset.
\\FMNIST dataset comprises 70000 grey-scale images that are equally distributed between 10 classes. This dataset is considered one of the benchmark datasets in computer vision and as a result, we decide to use it in our research. We resize the images from 28x28 to 24x24. 
\\We adopt the state of the art optimizer, Adam \cite{b2}, as our chosen optimizer and L2 regularization as the regularization technique. Furthermore, we employ exponential learning rate decay with a factor of 0.96 in the training process. we divide the model’s learning rate by 10 each time the accuracy of the model doesn’t improve after a number of epochs. For Yale face dataset, Belgium TS dataset, and FMNIST dataset, we train the models with batch sizes of 32, 16, and 128 respectively. You can find our implementation of experiments at \cite{b19}.

\subsection{Baseline Model}
Since our goal is to display that using different numbers of EM routing iteration in each of the stages can result in a more efficient way of training the CapsNets, we choose a CapsNet with d1=d2=d3=2 as our baseline method. This way of setting the iteration numbers is reported to be effective when training the CapsNets according to the previous works \cite{b9,b14,b15,b7}.

\subsection{Results from training the networks}
In the following, we compare the results from training the CapsNet with different numbers of EM routing iterations in terms of the highest accuracy they can achieve and also how efficiently they can be trained. Due to the limitations of time and resources, we are not able to try every possible combination of routing iterations. We start by training the baseline model then we change the number of iterations for each stage. Based on the performances of our models on each dataset, we select values 90\%, 70\% and 87\% as thresholds for test accuracy on Belgium TS dataset, Yale face dataset, and FMNIST dataset to measure how fast the models can perform. 

\subsubsection{Results on Belgium TS Dataset}
Results obtained from training our models on the Belgium TS dataset (Table \ref{Belgiumtab}) show that model 212 has outstanding results both in terms of speed and accuracy. This model surpasses the highest test accuracy achieved by our baseline model by 2.84\% while it is able to reach the threshold almost three times faster than the baseline and converges two times faster. For this dataset, the accuracy scored by model 212 is the highest accuracy achieved by CapsNets with EM routing in the literature \cite{b13}. Models 112 and 122 also achieve higher accuracies than the baseline model while both need less amount of time for reaching the threshold and converging. The fourth-place belongs to model 223 which achieves a test accuracy slightly higher than the baseline. However, this model needs more training compared to the baseline to achieve the threshold accuracy and, it takes more time to converge. Other models fail to achieve any results better than the baseline model. Model 221 takes a lot of time to converge and also, it fails to reach the accuracy threshold.

\subsubsection{Results on Yale Face Dataset}
Studying the results attained by training our models on the Yale face dataset (Table \ref{Yaletab}) demonstrates that the model 212 repeats its success on this dataset as well and achieves 11\% improvement compared to the baseline model and stands as the most accurate model. Furthermore, this model manages to reach the accuracy threshold two times faster than the baseline model. Model 112 also behaves the same as before and it keeps its place as the fastest model in reaching the threshold. The rest of the models are not able to make any improvements to the baseline results. One thing that catches the eye is that model 221, again, has a dramatically worse result compared to baseline and other models.

\subsubsection{Results on FMNIST dataset}
The results emerged from training our models on the FMNIST dataset (Table \ref{FMNISTtab}) displays that all the models can yield acceptable results. In agreement with our previous experiments, models 212 and 112 exhibit remarkable performances with an improvement of more than 2\% over the baseline results. Moreover, model 122 has managed to reach the accuracy threshold fastest and also converges sooner than other models. Like before, model 221 doesn't display any impressive achievement and only attains a poor accuracy of almost 18\%.

\begin{table*}[ht]
\caption{result from training moldes on belgium ts dataset}
\begin{center}
\begin{tabular}{|c|c|c|c|c|c|}
\hline
\textbf{}&\multicolumn{5}{|c|}{\textbf{Training Times and Test Accuracies}} \\
\cline{2-6}

\textbf{Models} & \textbf{\textit{Epochs}}& \textbf{\textit{Total time}}& \textbf{\textit{Best}}& \textbf{\textit{Best}}& \textbf{\textit{Total Time}}\\

\textbf{} & \textbf{\textit{until 90\%}}& \textbf{\textit{until 90\%(mins)}}& \textbf{\textit{accuracy}}& \textbf{\textit{epoch}}& \textbf{\textit{(mins)}}\\
\hline
Model222&73&161.33&92.13\%&109&240.89\\
Model212&30&57.6&\textbf{94.94\%}&74&142.08\\
Model112&30&\textbf{53.7}&92.71\%&72&128.88\\
Model122&55&107.25&93.22\%&89&173.55\\
Model232&151&401.66&91.03\%&166&441.56\\
Model113&85&165.75&90.62\%&85&165.75\\
Model223&100&247&92.28\%&140&345.8\\
Model322& & &88.68\%&87&215.76\\
Model221& & &59.81\%&162&336.96\\
\hline
\end{tabular}
\label{Belgiumtab}
\end{center}


\caption{result from training moldes on yale face dataset}
\begin{center}
\begin{tabular}{|c|c|c|c|c|c|}
\hline
\textbf{}&\multicolumn{5}{|c|}{\textbf{Training Times and Test Accuracies}} \\
\cline{2-6}

\textbf{Models} & \textbf{\textit{Epochs}}& \textbf{\textit{Total time}}& \textbf{\textit{Best}}& \textbf{\textit{Best}}& \textbf{\textit{Total Time}}\\

\textbf{} & \textbf{\textit{until 70\%}}& \textbf{\textit{until 70\%(mins)}}& \textbf{\textit{accuracy}}& \textbf{\textit{epoch}}& \textbf{\textit{(mins)}}\\
\hline
Model222&213&200.22&74.62\%&240&225.6\\
Model212&135&110.7&\textbf{85.94\%}&223&182.86\\
Model112&144&\textbf{103.68}&78.18\%&209&150.48\\
Model122&243&204.12&70.26\%&243&204.12\\
Model232& & &63.85\%&266&276.64\\
Model113& & &59.56\%&168&127.68\\
Model223& & &48.09\%&154&149.38\\
Model322& & &69.16\%&261&268.83\\
Model221& & &18.19\%&192&172.8\\
\hline
\end{tabular}
\label{Yaletab}
\end{center}


\caption{result from training moldes on fmnist dataset}
\begin{center}
\begin{tabular}{|c|c|c|c|c|c|}
\hline
\textbf{}&\multicolumn{5}{|c|}{\textbf{Training Times and Test Accuracies}} \\
\cline{2-6}

\textbf{Models} & \textbf{\textit{Epochs}}& \textbf{\textit{Total time}}& \textbf{\textit{Best}}& \textbf{\textit{Best}}& \textbf{\textit{Total Time}}\\

\textbf{} & \textbf{\textit{until 87\%}}& \textbf{\textit{until 87\%(mins)}}& \textbf{\textit{accuracy}}& \textbf{\textit{epoch}}& \textbf{\textit{(mins)}}\\
\hline
Model222&32&195.52&87.99\%&127&775.97\\
Model212&12&66.12&90.14\%&89&490.39\\
Model112&13&60.06&\textbf{90.63\%}&81&374.22\\
Model122&10&\textbf{53.1}&89.17\%&61&323.91\\
Model232&24&160.8&88.37\%&94&629.8\\
Model113&17&79.05&88.59\%&74&344.1\\
Model223&35&217.35&87.73\%&61&378.81\\
Model322&26&179.4&88.13\%&60&414\\
Model221& & &85.86\%&77&468.93\\
\hline
\end{tabular}
\label{FMNISTtab}
\end{center}
\end{table*}

\subsection{Training Summary}
Based on the results provided in Table \ref{Belgiumtab}, Table \ref{Yaletab}, and Table \ref{FMNISTtab}, we can observe that setting each of the d1, d2, and d3 from 2 to 3 iterations can cause the model to have a longer training time with no exciting results. On the other hand, we encounter different outcomes when setting the EM iteration numbers from 2 to 1. Models with d1=1 and/or d2=1 produces better or close results compared to the baseline model while they are trained faster, and decreasing the d3 leads to poorer performance compered to the baseline and other models.

\section{Conclusion and Recommendation}
In this work, we have tried to improve the performance of Matrix CapsNets by altering the number of iterations that the EM routing algorithm is run for the network. We have selected a baseline model that has been found to be a successful setting of CapsNets in the previous works in the literature. This model (model 222) uses two iterations of the EM routing algorithm in the last three stages of the network. Unlike the past studies, we have tried using un-equal numbers of EM routing iterations for different stages of the CapsNet. We have found the setting of d1=2, d2=1, and d3=2 for EM routing iterations to work very well. Using this setting, we have managed to achieve higher accuracies than the baseline model while training the model up to three times faster.
In conclusion, changing the number of iterations can have an immense effect on the performance of the CapsNets. Our research suggests using models 212, 112 and 122 instead of other models that utilize higher numbers of EM routing iterations (like 222 and 333) when attacking a new problem. Our proposed models can be trained fast with high accuracies; thus, you can decide whether the CapsNets are a plausible solution for your work.

\end{document}